\documentclass[acmtog, nonacm]{acmart}

\usepackage{booktabs} 

\usepackage[ruled]{algorithm2e} 

\SetAlFnt{\small}
\SetAlCapFnt{\small}
\SetAlCapNameFnt{\small}
\SetAlCapHSkip{0pt}

\definecolor{green}{rgb}{0, 0.7, 0}
\definecolor{deepgreen}{rgb}{0.0, 0.55, 0.2}

\newcommand{\zt}[1]{{#1}}
\newcommand{\rv}[1]{{#1}}

\newcommand{\cmt}[1]{}

\newcommand{\reffig}[1]{{Fig.~\ref{fig:#1}}}
\newcommand{\reftab}[1]{{Tab.~\ref{tab:#1}}}
\newcommand{\refsec}[1]{{Sec.~\ref{sec:#1}}}
\newcommand{\refeqn}[1]{{Eqn.~(\ref{eqn:#1})}}

\usepackage{makecell}
\usepackage[normalem]{ulem}
\usepackage[skip=0pt]{caption}

\begin{document}
\title{LVCD: Reference-based Lineart Video Colorization with Diffusion Models}

\author{Zhitong Huang}
\email{luckyhzt@gmail.com}
\affiliation{
    \institution{City University of Hong Kong}
    \city{Hong Kong SAR}
    \country{China}
}

\author{Mohan Zhang}
\email{zeromhzhang@tencent.com}
\affiliation{
    \institution{WeChat, Tencent Inc.}
    \city{Shenzhen}
    \country{China}
}

\author{Jing Liao}
\email{jingliao@cityu.edu.hk}
\affiliation{
    \institution{City University of Hong Kong}
    \city{Hong Kong SAR}
    \country{China}
}
\authornote{Corresponding author.}

\begin{abstract}
We propose the first video diffusion framework for reference-based lineart video colorization.
Unlike previous works that rely solely on image generative models to colorize lineart frame by frame, our approach leverages a large-scale pretrained video diffusion model to generate colorized animation videos.
This approach leads to more temporally consistent results and is better equipped to handle large motions.
Firstly, we introduce \textit{Sketch-guided ControlNet} which provides additional control to finetune an image-to-video diffusion model for controllable video synthesis, enabling the generation of animation videos conditioned on lineart.
We then propose \textit{Reference Attention} to facilitate the transfer of colors from the reference frame to other frames containing fast and expansive motions.
Finally, we present a novel scheme for sequential sampling, incorporating the \textit{Overlapped Blending Module} and \textit{Prev-Reference Attention}, to extend the video diffusion model beyond its original fixed-length limitation for long video colorization.
Both qualitative and quantitative results demonstrate that our method significantly outperforms state-of-the-art techniques in terms of frame and video quality, as well as temporal consistency.
Moreover, our method is capable of generating high-quality, long temporal-consistent animation videos with large motions, which is not achievable in previous works.
Our code and model are available at \textit{\textcolor{blue}{ \url{https://luckyhzt.github.io/lvcd}}}.
\end{abstract}



\begin{teaserfigure}
    \centering
    \includegraphics[width=1.0\textwidth]{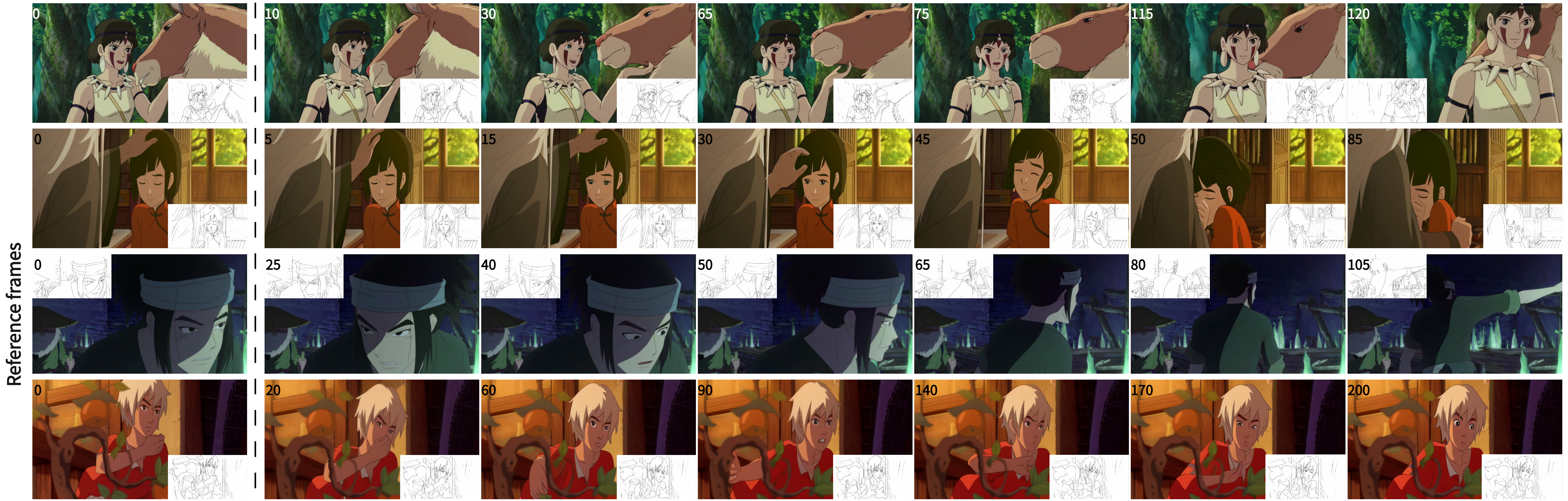}
    \caption{\zt{Given a reference image and a sequence of lineart, our method aims to colorize the sketches to produce long temporal-consistent animation videos. The upper-left number indicates the frame index. Input frames: $1^{st}$ row is from movie \textit{Princess Mononoke}, $2^{nd}$ and $4^{th}$ rows are from movie \textit{Big Fish \& Begonia}, $3^{rd}$ row is from movie \textit{Mr. Miao}.}}
    \label{fig:teaser}
\end{teaserfigure}

\settopmatter{printacmref=false}
\setcopyright{none}
\renewcommand\footnotetextcopyrightpermission[1]{}

\maketitle{}

\section{Introduction}
\label{sec:introduction}
Animation, as a prominent form of media and entertainment, continues to captivate audiences worldwide.
However, the production of animation remains a labor-intensive and time-consuming endeavor.
Traditionally, artists manually sketch and colorize keyframes, leaving the in-between frames to be completed in accordance with the style and color palette of the keyframes.
This filling process, while essential for preserving consistency and coherence in the animation, can be highly repetitive and tedious for the artists involved.
Therefore, an automated method for colorizing the in-between frames is crucial for boosting efficiency and productivity in the animation industry.

While previous works have attempted lineart video colorization, their frameworks typically rely on image generative models that process one frame at a time, focusing solely on the coherence between neighboring frames.
To maintain temporal consistency in the generated video, most of these works employ the previously generated frame as the new reference frame to produce subsequent frames.
Unfortunately, this propagation can lead to significant error accumulation and artifacts, even within 10 consecutive samplings (an example is provided in $2^{nd}$ and $3^{rd}$ rows of \reffig{compare}).
Another drawback of previous approaches is they all use frameworks based on Generative Adversarial Networks \cite{gan} (GANs), which have limited generative ability compared to more recent architectures such as transformers \cite{attention} and diffusion models \cite{ddpm}

Recently, generative models have garnered significant attention from researchers.
Among these, diffusion models \cite{ddpm} have emerged as particularly powerful tools for generating visually appealing content, including images and videos \cite{diff2gan, latent-video, animatediff}. 
Leveraging the strengths of diffusion models, we propose the first video diffusion framework for lineart video colorization based on reference frame.
Following extensive evaluation, we select Stable Video Diffusion \cite{svd} (SVD) as the base model for our framework.
This decision is driven by two key factors: 
a) the large amount of training data empowers SVD with robust generative capabilities to \rv{synthesize} high-fidelity videos; 
b) the temporal video model facilitates the generation of temporally consistent videos, surpassing image generative models.

While SVD provides a robust foundation for our proposed task, we must address three significant challenges to adapt it accordingly:
1). \textit{Additional control of lineart sketch}: The original SVD, as an image-to-video model, only supports the conditioning from a reference image. Introducing additional control of the lineart sketch is essential to the model's adaptation.
2). \textit{Adaptation to expansive motions}: While SVD is limited to produce videos with subtle motions, animation clips often feature larger motions. Thus, modifications to the original SVD are necessary to accommodate expansive motions in animations.
3). \textit{Extension to long video}: The original SVD is restricted to generating videos of fixed length, which does not meet the requirements for animation colorization. Therefore, extending the original SVD to generate long, temporally consistent videos is imperative.

We tackle the aforementioned challenges by introducing the first reference-based lineart video colorization framework with a diffusion model. 
This framework is adept at generating high-quality, large-motion, and temporally consistent animation videos guided by lineart sketches.
To achieve this, we first extend the ControlNet \cite{controlnet} to a video version, termed \textit{Sketch-guided ControlNet}, incorporating additional lineart sketch control as a crucial guide for the animation's layout and structure.
Secondly, we introduce \textit{Reference Attention} to replace the original spatial attention layers in SVD, facilitating long-range spatial matching between the first reference frame and consecutive generated frames. 
This modification enhances the model's ability to colorize frames with large motions relative to the reference frame. 
Lastly, in contrast to the original SVD, which is trained to generate videos of fixed length (e.g., 14 frames) and thus inadequate for generating long videos with temporal consistency, we introduce a novel scheme for sequential sampling, including \textit{Overlapped Blending Module} and \textit{Prev-Reference Attention}, to enable the colorization of long animation.

The extensive experiments demonstrate the effectiveness of our method in generating long, temporally consistent animations with large motions of high fidelity, given a reference frame and a sequence of lineart. In summary, our contributions are three-fold:
\begin{itemize}
    \item We propose the first video diffusion framework for reference-based lineart animation colorization, harnessing the capabilities of a pretrained video diffusion model to generate long, temporally consistent animations of high quality.
    \item We introduce Reference Attention for SVD, enhancing the model's ability to generate animations with swift motions.
    \item We design a novel sequential sampling mechanism, including the Overlapped Blending Module and Prev-Reference Attention, to extend the model to produce long animations with long-term temporal consistency.
\end{itemize}

\vspace{-7pt}
\section{Related Works}
\label{sec:related}
\vspace{3pt} \noindent \textbf{Lineart Image Colorization. \;}
\rv{Unlike} natural image colorization, where the grayscale image retains an illuminance channel for colors, lineart exclusively contains structural information.
This characteristic affords greater freedom and flexibility in lineart colorization.
Traditional optimization-based methods for lineart colorization \cite{manga_color, lazy_brush} typically involve users manually brushing desired colors onto specific regions.
The emergence of deep neural networks has propelled lineart colorization into a new realm, where various techniques have been employed, including color hint points \cite{two_stage_hint}, color scribbles \cite{user_guide_cgan}, text tags \cite{tag2pix}, and even natural language \cite{language_color}.
Another significant approach is reference-based lineart colorization, wherein users \rv{only need to provide} a reference cartoon image to colorize the lineart with similar colors and styles.
The crux of this method lies in establishing correspondence between the lineart and the reference image. 
Sato et al. \cite{graph_correspond} proposed a matching technique based on segmented graph structures, while Chen et al. \cite{active_color} introduced an active learning framework for matching, and Yu et al. \cite{attn_color} employed an attention-based network. 
The first diffusion-based framework for anime face lineart colorization, based on reference images, was proposed in \cite{anime_diffusion}. 
Despite the advancements in lineart image colorization, these frameworks have limitations in producing videos without accounting for temporal coherence between colorized frames.

\vspace{3pt} \noindent \textbf{Reference-based Lineart Video Colorization. \;}
Compared to lineart image colorization, which involves various types of user interactions and conditions, reference frame is the most significant and reasonable approach for lineart video colorization, which can be directly reused for subsequent frames without human intervention.
Several methods extend reference-based colorization to lineart videos.
Zhang et al. \cite{corr_match} trained a reference-based lineart image colorization framework using frame pairs from animation video clips.
However, this method only preserves consistency with the reference frame, lacking temporal coherence between the generated frames.
In \cite{temporal_video_color}, previously colorized frames are used as new reference frames for subsequent frame generation to maintain short-term temporal coherence.
Nevertheless, this may lead to error accumulation as the video length increases.
Wang et al. \cite{transform_anime_color} addressed this by using both the first reference frame and previously generated frames as references, potentially mitigating error accumulation.
Yu et al. \cite{flow_anime_color} proposed propagating colors from the reference frame to subsequent frames based on estimated optical flow, but refinement is still required afterward, leading to certain degree of error accumulation.
In \cite{artist_guide_color}, a user-guided framework preserved temporal coherence by requiring users to provide dense and consistent color hint points.
However, this method contradicts the initial motivation of automatic lineart video colorization due to its significant demand for user interaction and proficiency.
All aforementioned works rely on image generative models, limiting the model's ability to produce long, temporally consistent animations. 
Our framework, based on the video diffusion model, addresses this limitation by generating long, temporally consistent animations through sequential sampling.

\zt{
\vspace{3pt} \noindent \textbf{Video Interpolation. \;}
Unlike reference-based lineart video colorization, where only the first frame is provided, video interpolation frameworks aim to predict in-between frames given both the first and last frames.
Building on a pretrained text-to-video diffusion model, SparseCtrl \cite{sparsectrl} introduced additional image conditions to perform video interpolation.
SEINE \cite{seine} proposed a diffusion-based framework for video transitions using masked frame conditions, allowing for differences in content and style between the first and last frames.
The concept of video interpolation is further adapted to cartoon animations, where a temporal constraint network is incorporated in \cite{few_reference_color}.
Li et al. \cite{cartoon_inbetween} developed a framework for colorizing in-between lineart sketches based on optical flow.
AnimeInterp \cite{anime-interp} interpolates a single middle frame through warping with predicted flows from segment-guided matching.
To improve the perpetual quality of animation interpolation, EISAI \cite{eisai} proposed a forward-warping interpolation architecture to prevent line destruction and ghosting artifacts.
More recently, ToonCrafter \cite{tooncrafter} introduced a sketch-guided video diffusion model for animation interpolation.
Despite these advancements, the task of animation interpolation remains limited to generating a fixed number of intermediate frames, requiring users to provide extensive reference frames for colorizing long sequences of lineart sketches.
}

\vspace{3pt} \noindent \textbf{Video Diffusion Models. \;}
As advancements in diffusion-based image synthesis \cite{diff-beat-gan, custom-diff, diff2gan} continue, numerous diffusion-based frameworks for video synthesis have emerged. 
Some approaches involve training a video diffusion model with temporal layers from scratch \cite{latent-video, trip}, while a more common method is to add temporal layers to pretrained image diffusion models and finetune them for video synthesis \cite{latte, animatediff, nuwa-xl}.
SVD \cite{svd} is one such framework, initiated from an image latent diffusion model \cite{ldm}, and adapted to video synthesis with additional temporal layers including 3D convolution and temporal attention.
\zt{In \cite{control-a-video}, structural controls, e.g., canny, depth, and HED, are further \rv{incorporated} to guide the video generation by \rv{introducing} ControlNet \cite{controlnet} architecture to video diffusion models.}
Despite the existence of video diffusion models for generating both realistic videos \cite{latte} and animations \cite{animatediff}, \rv{none performs} lineart video colorization \rv{using} a reference frame.
Building upon SVD, a large-scale pretrained video diffusion model, we introduce the first diffusion-based framework tailored for this specific task.

\section{Preliminary}

We introduce the preliminary of Stable Video Diffusion \cite{svd} (SVD), the backbone image-to-video model for our work, including the diffusion framework (EDM \cite{edm}) and Euler-step sampling utilized in SVD.

\begin{figure*}[t]
	\centerline{\includegraphics[width=1.0\textwidth]{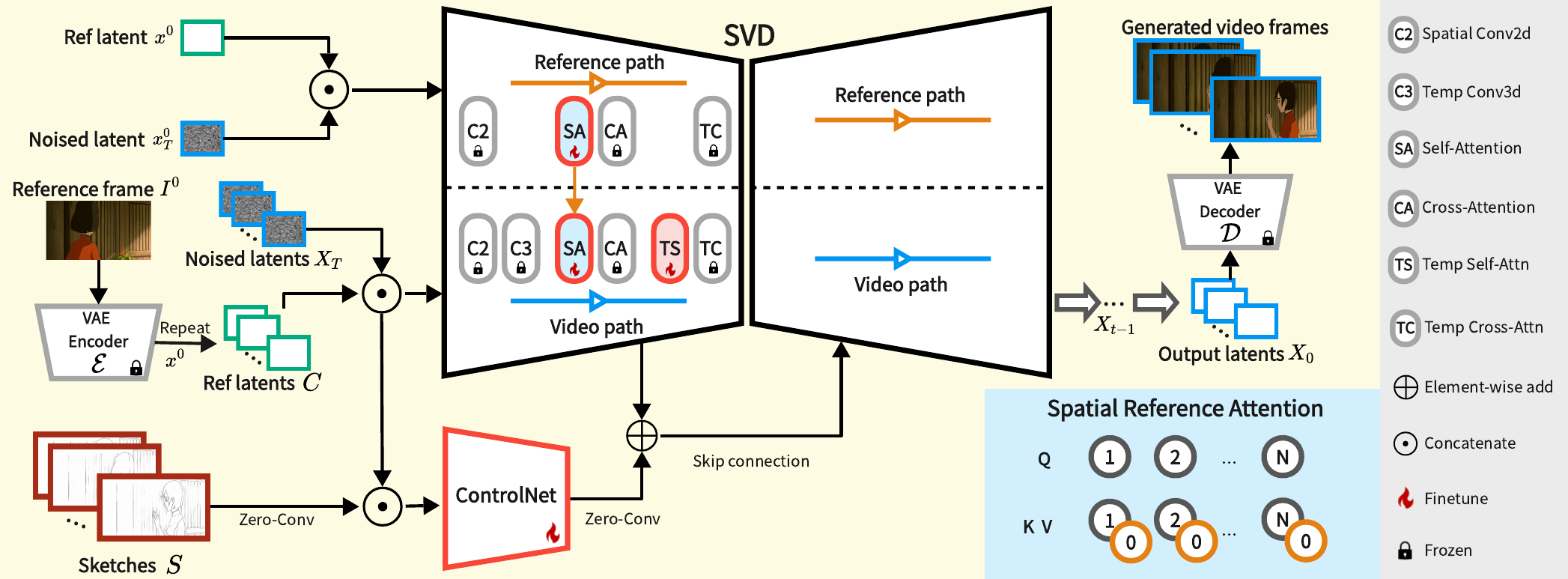}}
	\caption{\zt{Model architecture of sketch-guided ControlNet and Reference Attention. All frames are from \textit{Big Fish \& Begonia}.} \cmt{The input latents $X_T$ are concatenated with the encoded conditional latents $C$ and then fed to the UNet. For the ControlNet branch, the inputs are further concatenated with the lineart sketches $S$. We make modifications to all the spatial self-attention and temporal self-attention layers to add reference and mask respectively. Accordingly, we only finetune these two types of layers in the UNet together with the ContrlNet, and freeze other layers in the UNet.}}
	\label{fig:architecture}
\end{figure*}

\vspace{3pt} \noindent \textbf{Stable Video Diffusion. \;}
We adopt SVD as our backbone model, composed of a Variational Autoencoder \cite{vae} (VAE) and a U-Net \cite{unet}, illustrated in \reffig{architecture}.
The VAE encoder maps input video frames into a lower-dimensional latent space, and the VAE decoder reconstructs the latents back into frames, with temporal layers introduced to mitigate minor flickering artifacts.
Then, the U-Net, initialized from the text-to-image Stable Diffusion \cite{ldm} (SD), is finetuned to denoise fixed-length sequences of latents (i.e., 14 frames), incorporating additional temporal attention and 3D convolutional layers.
Originally, SVD was trained for the image-to-video task, where the first frame is provided to guide the denoising process.

\vspace{3pt} \noindent \textbf{EDM. \;}
In SVD, the denoising network is trained with EDM, a continuous-time diffusion framework. The denoiser $D_\theta$ is trained with denoising score-matching (DSM) loss, given by:
\begin{equation}
\label{eqn:edm_loss}
    \mathbb{E}_{(x_0,c) \sim p_{data} (x_0, c), (\sigma, n) \sim p(\sigma, n)}
    \Bigl[ \lambda_\sigma  \left\| D_\theta(x_0+n;\sigma,c) - x_0 \right\|^2_2  \Bigr]
\end{equation}
where $p(\sigma, n)$ is the distribution of noise level $\sigma$ and normal noise $n$, $\lambda_\sigma$ is a loss weighting function for different noise levels, and $c$ represents the conditioning signal (e.g., conditional frame in SVD).
The denoiser $D_\theta$ receives the clean image from the outputs of the UNet $U_\theta$:
\begin{equation}
\label{eqn:denoiser}
    D_\theta(x;\sigma, c) = c_{skip}(\sigma) \cdot x + c_{out}(\sigma) \cdot U_\theta \Bigl( c_{in}(\sigma) \cdot x;c_{noise}(\sigma), c \Bigr)
\end{equation}
where $c_{skip}(\sigma)$, $c_{out}(\sigma)$, $c_{in}(\sigma)$ and $c_{noise}(\sigma)$ are EDM preconditioning parameters dependent on noise level.
We adhere to the same training scheme used in SVD to finetune the model.

\vspace{3pt} \noindent \textbf{Euler-step Sampling. \;} During sampling, we employ Euler-step to gradually obtain the clean image $x_0$ from Gaussian noise $x_T$ as in SVD:
\begin{equation}
\label{eqn:euler-step}
    x_{t-1} = \frac{\sigma_{t-1}}{\sigma_t} x_t + \frac{\sigma_t - \sigma_{t-1}}{\sigma_t} D_\theta(x_t; \sigma_t, c)
\end{equation}
where $\sigma_t$ represents the discretized noise level for each timestep $t \in \left[ 0, T \right]$.

\section{Methodology}
\label{sec:methodology}

We aim to design a video diffusion framework for reference-based lineart video colorization, capable of producing temporally consistent long sequences of animations with large motions.
First, in \refsec{architecture}, we discuss the model architecture, including the sketch-guided ControlNet and Reference Attention, which enable the model to generate animations with fast and expansive movements guided by lineart sketches.
After modifying the model architecture, we finetune it using animation videos to perform our task.
During inference, we extend the original SVD to produce long, temporally consistent animations through sequential sampling, incorporating the Overlapped Blending Module and Prev-Reference Attention, as discussed in \refsec{sample-scheme}.

\subsection{Model Architecture}
\label{sec:architecture}
An overview of our framework is shown in \reffig{architecture}. Our objective is to colorize a sequence of lineart sketches $S=s^{[1:N]}=\{ s^1,...,s^N \}$ given a reference image $I^0$, to produce a sequence of video frames $I^{[1:N]}$ (currently we assume the video length equals to the input length $N$ of the model) with a diffusion-based framework:
\begin{equation}
\label{eqn:sample}
    X_0 = D_\theta ( X_T, S, C )
\end{equation}
where $D_\theta$ is the denoiser, $X_T = x_T^{[1:N]} = \{ x_T^1, ..., x_T^N \}$ is the initial noised latents sampled from Gaussian distribution, $C = [x^0] \times N$ is the conditional input of repeated reference latents which are concatenated to noised latents forming the $i^{th}$ input as $[x^0, x_T^i]$, where $i = 1,..., N$. Here, $x^0 = \mathcal{E} (I^0)$ is the encoded reference frame. For simplicity, we omit the Euler-step sampling and directly obtain the denoised video latents $X_0$, which are then decoded into video frames $I^{[1:N]} = \mathcal{D}(x_0^{[1:N]})$.

\vspace{3pt} \noindent \textbf{Sketch-guided ControlNet. \;}
Besides the reference image, another key condition is the lineart sketch, which is not supported in the original SVD. As shown in \reffig{architecture}, we adapt the design from ControlNet \cite{controlnet} to incorporate the sketch as an additional condition.
Firstly, we duplicate the original UNet's encoder, cloning all the layers, including temporal attention and 3D convolutional layers, along with their weights.
Secondly, we introduce several zero-initialized convolutional layers to encode the lineart sketches and concatenate them to the input of the cloned encoder.
Finally, the outputs of each layer of the ControlNet are added to the skip connections of the original U-Net decoder.
During training, all the layers in ControlNet are finetuned to generate video sequences conditioned on both the reference image and the lineart sketches.


\vspace{3pt} \noindent \textbf{Reference Attention. \;}
Animation videos often feature fast and large motions due to their low frame rate and virtual nature, necessitating the model to perform long-range spatial matching between the reference frame and the frames to be colorized.
However, the original video path of SVD struggles to meet these requirements for two main reasons.
Firstly, the encoded reference latents are concatenated with the noised latents along the channel dimension to form the video input $[x^0, x_T^i], i = 1, ..., N$, as shown in \reffig{architecture}.
This spatial alignment of the two inputs prevents effective long-range matching between them.
Secondly, the temporal layers in the video path which \rv{interact} between the $N$ inputs are one-dimensional, considering only the correlation between frames at the same spatial position, thereby neglecting the relationship between frames at different positions.
These two factors limit the model's ability to propagate color information effectively from the reference frame to distant frames with large motions.

As shown in \reffig{architecture}, to enable long-range spatial matching, we propose to calculate global attention between reference latent and noised frame latent. 
We concatenate the encoded reference latent with a noised latent to form the reference input $[x^0, x_T^0]$ and feed it to a new reference path, where all temporal layers (temporal Conv3d and temporal self-attention layers) are skipped.
We then insert the intermediate features of the reference path to interact with the features of the video path within the spatial self-attention layers using Reference Attention:
\begin{equation}
\label{eqn:refattn}
    \mathrm{Attn}(Q_i, K_i, V_i) \rightarrow \mathrm{Attn}(Q_i, \left[ K_i, K_0' \right], \left[ V_i, V_0' \right] )
\end{equation}
where $\left[ .,. \right]$ represents spatial concatenation, $K_0'$ and $V_0'$ are the key and value mappings for intermediate features of the reference input $[x^0, x_T^0]$ fed into the reference path, and $i$ is the index of the $N$ inputs going through \rv{the} video path.
As shown in \reffig{architecture}-Spatial Reference Attention, each frame performs attention not only with itself but also queries information from the reference input (labeled as orange blocks).
Since the SVD model is trained to reconstruct the ground-truth latent $x^i$ for each input $[x^0, x_t^i]$ using the \refeqn{edm_loss}, the input $[x^0, x_t^0]$ to the reference path has also learned to reconstruct $x^0$, thereby ensuring that the intermediate features of the reference path contain complete information about the reference latent $x^0$.
This modified Reference Attention calculates the global attention between the reference frame and the frames, enabling effective propagation of content to distant positions when large motions are engaged.

Finally, we use the loss in \refeqn{edm_loss} to finetune the modified network with sketch-guided ControlNet and Reference Attention. We update all the layers in ControlNet as well as the spatial and temporal self-attention layers in UNet, as illustrated in \reffig{architecture}.

\subsection{Sequential Sampling for Long Animation}
\label{sec:sample-scheme}

\begin{figure*}[t]
    \centerline{\includegraphics[width=1.0\linewidth]{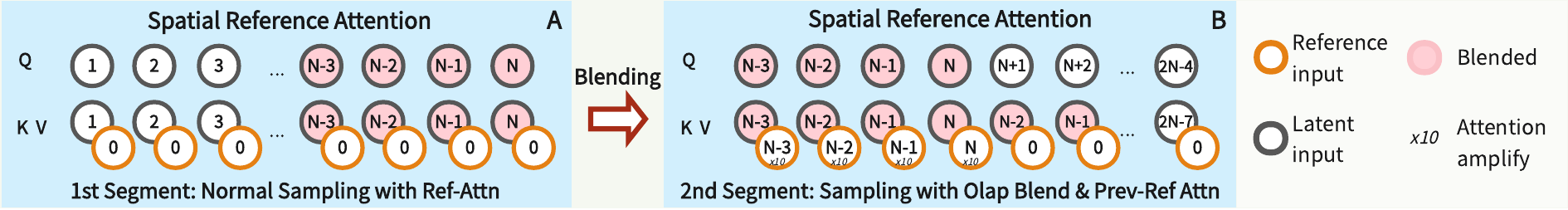}}
    \caption{Sequential sampling with Overlapped Blending and Prev-Reference Attention.}
    \label{fig:sample}
\end{figure*}

With sketch-guided ControlNet and Reference Attention, our model can already generate high-quality animations of a fixed length $N$.
For longer video sequences, we divide the video into segments and generate each segment sequentially.
To avoid accumulated errors, we consistently use the first reference frame $I^0$ for samplings of all segments.
However, this sampling scheme only ensures temporal consistency within the same segment.
To introduce additional dependencies between segments, we utilize a certain number of overlapped frames.
Specifically, we divide the video sequence $I^{[1:L]}$ of length $L$ into segments indexed from $1$ to $\frac{L-o}{N-o}$, where $o$ is the number of overlapped frames.
For instance, the $n^{th}$ segment is $V^n = I^{[(n-1)(N-o)+1:(n-1)(N-o)+N]}$.
Based on the scheme of sequential sampling, we then introduce Overlapped Blending Module and Prev-Reference Attention to preserve the long-term temporal coherence across consecutive segments.



\vspace{3pt} \noindent \textbf{Overlapped Blending Module. \;}
After splitting the video into overlapped segments, the beginning $o$ frames of the segment $V^n$, are already generated in the preceding segment.
To incorporate this information into subsequent segments, we use two methods.
Firstly, we extend spatial blending \cite{blending}, originally used for image inpainting, to temporal blending.
As shown in \reffig{sample}, for the overlapped frames (highlighted in pink), we replace the intermediate denoising results of segment $V^{n}$ with those from the previous segment $V^{n-1}$ for all denoising timesteps $t$:
\begin{equation}
    D_\theta ( X_t^{n}, S^{n}, C^{n}, R ) \leftarrow D_\theta ( X_t^{n-1}, S^{n-1}, C^{n-1}, R )
\end{equation}

Secondly, we further insert the contents of the previously generated frames via Reference Attention.
As shown in \reffig{sample}-B, we first feed all previously generated results of the overlapped frames (in orange blocks) as reference input to the reference path as mentioned in \refsec{architecture}, which captures complete information from these inputs.
Then, the spatial Reference Attention of the overlapped frames is defined as:
\begin{equation}
    \mathrm{Attn}(Q_i, \left[ K_i, \log{\alpha} \cdot K_i' \right], \left[ V_i, V_i' \right] ) , \:\: i \le o
\end{equation} 
where $K_i'$ and $V_i'$ are the mapped key and value of the previous result $x^i$ fed through the reference path with input $[x^i, x_T^i]$.
The term $\log{\alpha}$ is an amplifying factor that amplifies the attention weights of previous results by $\alpha = 10.0$ times.

\vspace{3pt} \noindent \textbf{Prev-Reference Attention. \;}
To effectively propagate the contents of the blended overlapped frames to distant frames, we further propose enhancing temporal propagation within the spatial self-attention layers.
As shown in \reffig{sample}-B, we shift the original self-attention of non-overlapped frames to the left by three frames: 
\begin{equation}  
\label{eqn:prevref}
         \mathrm{Attn}(Q_i, \left[ K_{i-3}, K_0' \right], \left[ V_{i-3}, V_0' \right] ), \:\: i > o
\end{equation}
By enabling the non-overlapped frames to query information from the overlapped frames, whose contents are restored to previously generated results through the Overlapped Blending Module, we effectively preserve the content consistency across consecutive segments.


By incorporating Overlapped Blending Module and Prev-Reference Attention, we guarantee long-term temporal coherence for extended long video synthesis.

\cmt{
\begin{algorithm}[t]
\caption{Sampling Algorithm \todo{may be removed} }
\SetAlgoNoLine
\KwIn{Denoiser $D_\theta$, 1st reference frame $I^r$, sketches $s^{[r, 1:L]}$, sample step $T$, input length $N$,  overlapping $o$. }
\KwOut{The colorized animation video sequence $I^{[1:L]}$}
Encode reference frame: $c = \mathcal{E} (I^r)$ and repeat: $C = [c] \times N$ \;
Initialize blending cache $H = h_{[1:T]}$ \;
\For{$n \gets 1$ \KwTo $(L-o) /(N-o)$}{
        Index: $i_0=(n-1)(N-o)+1$, $i_1=(n-1)(N-o)+N$ \;
        Sample initial noise: $X_T = x_T^{[1:N]} \gets \mathcal{N} (0, 1)$ \;
        Input sketches: $S \gets s^{[i_0:i_1]}$ \;
        
        $R = \left\{  
         \begin{array}{lr}  
         I^r, & n = 1 \\
         I^{[r, i_0:i_0+o]}, & n > 1
         \end{array} \right. $  \tcp{Input for Ref Attn}

        \For{$t \gets 1$ \KwTo $T$}{
            $X_0' \gets D_\theta ( X_t, S, C, R)$ \;

            \uIf{$n > 1$}{
                $ {x_0'}^{[1:o]} \gets h_t $ \;
            }
            
            $h_t \gets {x_0'}^{[-o:-1]} $ \;
            
            $X_{t-1} = \frac{\sigma_{t-1}}{\sigma_t} X_t + \frac{\sigma_t - \sigma_{t-1}}{\sigma_t} X_0'$  \;
        }
        $I^{[i_0:i_1]} \gets \mathcal{D}(X_0) $ \;
    }
    \label{alg:sample}
\end{algorithm}
}

\section{Implementation Details}
\vspace{3pt} \noindent \textbf{Dataset. \;}
We train our model with six animation movies directed by Hayao Miyazaki, comprising \textit{My Neighbor Totoro, The Secret World of Arriey, Whisper of the Heart, Ponyo, Kiki's Delivery Service}, and \textit{The Wind Rises}. 
\zt{For each movie, we first quantize the RGB values into 1,000 colors, then apply a threshold of $30.0$ on the bin-wise RMSE between the color histograms of consecutive frames to detect scene switches.}
We further exclude video clips with lengths smaller than 15 frames (the minimum required input length) and larger than 200 frames (which typically indicates an undetected scene switch).
The final filtered training set consists of 6,472 video clips with an average length of 58 frames. 
We then utilize the method in \cite{draw_sketch} to extract the lineart sketch for each frame.

\vspace{3pt} \noindent \textbf{Training. \;}
\rv{During training, we input $N+1$ frames per batch, where $N=14$ corresponds to the length of the original SVD input sequence, with an additional frame for Reference Attention.
For each video clip with frames $\{ I^1, ..., I^L \}$ of length $L$ in the training dataset, we split it into $L-N-1$ training sets indexed from $1$ to $L-N$, where the $k^{th}$ set consists of a sequence of consecutive frames $\{ I^{k+1}, ..., I^{k+N} \}$ and a group of candidate reference frames $\{ I^1, ..., I^{k} \}$.
For each training step, a reference frame is randomly selected from the candidate set to predict the $N$ consecutive frames, allowing for variable and random distances between the reference and predicted frames, which adapts to inference scenarios.}

\rv{We finetune the network using the loss function in \refeqn{edm_loss}, updating all layers in ControlNet and the spatial and temporal self-attention layers in UNet. 
During training, the noise level $\sigma$ is sampled from $\log \sigma = \mathcal{N} (1.0, 1.6)$. 
Due to limited computing resources, we train at a resolution of $576 \times 320$, as the original SVD is pre-trained with a resolution of $576 \times 320$ and then finetuned to $1024 \times 576$.
We use a batch size of 32 with a learning rate of $5.0\times10^{-5}$, and train for 18,000 steps over 80 hours on four NVIDIA RTX A6000 GPUs.}

\vspace{3pt} \noindent \textbf{Sampling. \;}
During sampling, we utilize $o=4$ overlapped frames. 
For the noise scale, we set $\sigma_{min}=0.002$ and $\sigma_{max}=700$, and then linearly decrease $\sigma_t^{1/\rho}$ from $\sigma_{max}^{1/\rho}$ to $\sigma_{min}^{1/\rho}$ as $t$ decreases from $T$ to $1$, where $\rho=7$. 
The denoising is performed for $T=25$ steps using the Euler step specified in \refeqn{euler-step}.
We apply Overlapped Blending and Prev-Reference Attention for all denoising steps to enhance temporal consistency.
The average inference time is approximately 2 seconds per frame on a single NVIDIA RTX A6000 GPU.

\section{Experimental Results}
\label{sec:results}

\subsection{Experimental Setup}
\label{sec:exp-setup}
\vspace{3pt} \noindent \textbf{Test Dataset. \;}
We choose four movies directed by Hayao Miyazaki, namely \textit{Howl's Moving Castle, Porco Rosso, Princess Mononoke}, and \textit{Spirited Away}, for our test set, labeled as \textit{Similar Testset} which exhibit similar artistic styles but different content from the training dataset.
Additionally, to assess the generalization ability of our model across diverse animation styles and contents, we select three movies, including \textit{Big Fish \& Begonia, Mr. Miao}, and \textit{LuoXiaoHei}, produced by other directors, denoted as \textit{General Testset}.
We segment the videos into clips and extract the lineart sketches using the same methods as employed for the training dataset.
\zt{For each testset, we evenly select 1,000 video clips for evaluation, with an average length of 59 frames.}
\zt{We measure the average movements of the two testsets with optical flow in resolution $256\times256$, where static positions are excluded. We found that 55\% of the clips have average motions larger than 5 pixels and 28\% of them exceed 10 pixels.}

\vspace{3pt} \noindent \textbf{Evaluation Metrics. \;}
We evaluate the quality of the colorized animations across four aspects:
\textit{1). Frame \& Video Quality}: We use \textbf{FID} \cite{fid} and \textbf{FVD} \cite{fvd} to evaluate the frame and video quality of the generated videos, respectively.
\textit{2). Frame Similarity}: Since the animation is generated conditioned on the lineart sketches and the first reference frame, both extracted from the original animation, we measure the similarity between the generated frames and the original animation frames using \textbf{PSNR}, \textbf{LPIPS}, and \textbf{SSIM} \cite{ssim}.
\textit{3). Sketch Alignment}: To evaluate whether the generated frames align with the structure of the input lineart sketches, we extract the sketches of the generated frames and calculate the Euclidean Distance Map \cite{ED-map} (ED Map), which measures the distances from each pixel to its nearest sketch. Subsequently, we compute the Euclidean Distance Map Difference (EDMD) in terms of RMSE, which indicates the average pixel shifts compared to the input sketches.
\textit{4). Temporal Consistency}:  We define the Temporal Consistency (\textbf{TC}) as:
\begin{equation}
    \mathrm{TC} = \frac{\lVert I_g^{t \rightarrow t+1} - I_g^{t+1} \rVert_2 \: / \: \lVert I_g^{t+1} \rVert_2}{\lVert I^{t \rightarrow t+1} - I^{t+1} \rVert_2 \: / \: \lVert I^{t+1} \rVert_2}
\end{equation}
where $I_g^t$ is the $t^{th}$ frame in the generated video, $I^t$ denotes the original frame, and $I^{t \rightarrow t+1}$ represents the warped frame $t+1$ from frame $t$.
Here, optical flow predicted from the original animation with RAFT \cite{raft} is utilized for both warpings in the original and generated frames.
For all the metrics, we resize the frames to $256 \times 256$ and normalize the pixel value to $[0.0, 1.0]$ for calculation.

\subsection{Comparisons on Reference-based Works}
We compare our proposed method with two existing reference-based lineart video colorization frameworks: ACOF \cite{flow_anime_color} (an optical-flow-based method) and TCVC \cite{temporal_video_color} (an image-to-image framework), both of which are GAN-based image model.
Since there is no widely accepted benchmark dataset for lineart video colorization, to ensure a fair comparison, we utilize our dataset to generate frame pairs for training ACOF and TCVC using their official code.
For both methods, we evaluate two versions: the original version, Prev Sample, updates the reference frame to the previously generated frame, while the modified version, First Sample, consistently uses the first frame as the reference.
Given the absence of diffusion-based frameworks for our task, \zt{we employ an image ControlNet \cite{controlnet} with AnythingV3 \cite{anything-v3}, finetuned from Stable Image Diffusion cartoon images and lineart controls.}
Additionally, we utilize Reference-only \cite{refonly} during sampling to \rv{provide reference frames as guidance}.
\zt{
We further select an animation interpolation work EISAI \cite{eisai} and a diffusion-based video interpolation work SEINE \cite{seine} for comparison.
}
\rv{Since both methods interpolate between colorized reference keyframes at a regular interval, we first apply the ControlNet + Reference-only method to colorize keyframes for every 13 frames before applying EISAI and SEINE to interpolate the remaining frames.}

\begin{figure*}[!ht]
    \centerline{\includegraphics[width=1.0\textwidth]{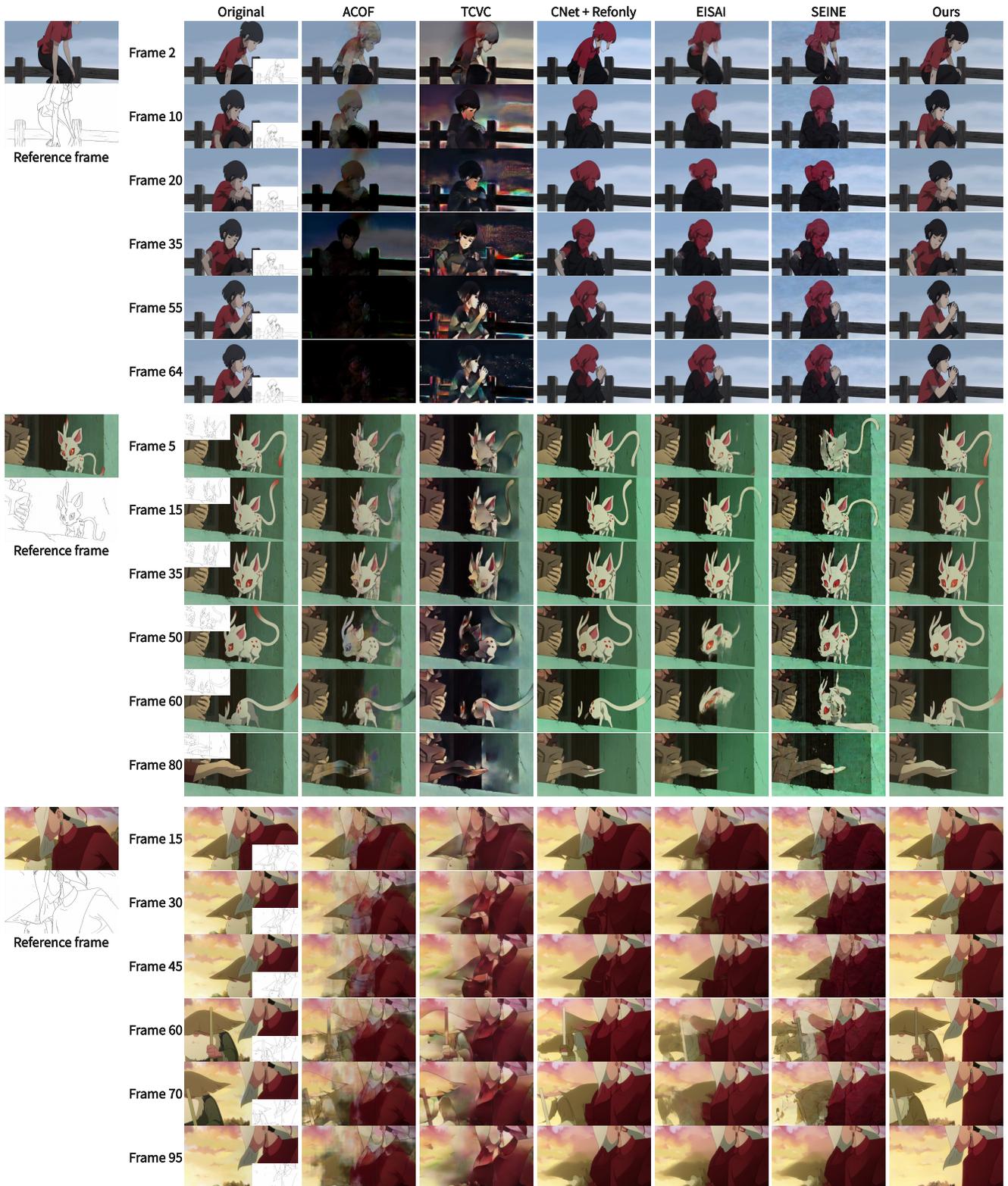}}
    \caption{\zt{Qualitative comparison with five methods: ACOF \cite{flow_anime_color}, TCVC \cite{temporal_video_color}, CNet+Refonly \cite{controlnet}, EISAI \cite{eisai}, and SEINE \cite{seine}. \rv{It is recommended to zoom in on the figure to observe the differences.} Input frames: $1^{st}$ and $2^{nd}$ examples are from \textit{Big Fish \& Begonia}, $3^{rd}$ example is from \textit{Mr. Miao}.}}
    \vspace{-10pt}
    \label{fig:compare}
\end{figure*}

\vspace{3pt} \noindent \textbf{Qualitative Comparison. \;}
In \reffig{compare}, we present colorized frames from three animation clips.
In the $1^{st}$ example, both ACOF and TCVC exhibit severe accumulated artifacts when using the original version of Prev Sample. 
In the $2^{nd}$ and $3^{rd}$ examples, even with First Sample, both methods struggle to generate frames with significant movements compared to the reference frame, resulting in a ghosting effect at the moving positions.
Hence, both methods are prone to certain artifacts, with either Prev or First Sample. 
This suggests that previous CNN-based methods struggle to effectively colorize long sequences of lineart sketches with substantial motions.
Regarding the image ControlNet plus Reference-only methods, they can generate frames successfully only when the movement range is limited.
The Reference-only technique, applied solely during inference and not trained with the model, may incorrectly interpret the correspondence from the reference frame when the object displacement is too large, as seen in the red hair of the girl in the $1^{st}$ example and \zt{the distorted elder in Frame 60 and 70 of the $3^{rd}$ example}.
The experiment highlights the limitations of the image diffusion model in fulfilling our task.
\zt{
For interpolation methods, EISAI exhibits ghosting effects, while SEINE generates significant noise which blurs the frames \rv{(zoom-in on the figure is suggested to note the artifacts)}. This suggests that neither method can be adapted to our task, even provided with keyframes generated by the CNet + Refonly method.
}


Our method, leveraging a video diffusion model with long-range spatial matching through Reference Attention, is capable of producing long temporal-consistent animations featuring large motions.
For instance, in all the examples, when the \rv{sprite and the characters} change positions, our method accurately locates the correct correspondence and effectively colorizes them.
Moreover, with the sequential sampling incorporating Overlapped Blending and Prev-Reference Attention, our method can preserve long-term temporal consistency.
\zt{As evidenced by the tail of the \rv{sprite} in the $2^{nd}$ example and the generated head (not in the reference frame) in the $1^{st}$ example}, a similar color is preserved throughout the entire animation.
Overall, our method successfully achieves the task of colorizing long sequences of linearts, a feat unattainable by previous works, including both CNN-based and diffusion-based frameworks.

\begin{table*}[t]
\setlength\tabcolsep{3.1pt} 
\footnotesize
\caption{\zt{Quantitative comparison with ACOF \cite{flow_anime_color}, TCVC \cite{temporal_video_color}, ControlNet \cite{controlnet} with Reference-only, and video interpolation methods EISAI \cite{eisai} and SEINE \cite{seine} }.}
\label{tab:comparison}
\begin{center}
\begin{tabular}{cccccccccccccccc}
    \toprule
    {} &  \multicolumn{7}{c}{Similar Testset}& {} &    \multicolumn{7}{c}{General Testset}\\
    {} & FID$\downarrow$ &    FVD$\downarrow$&LPIPS$\downarrow$&PSNR$\uparrow$&SSIM$\uparrow$& EDMD$\downarrow$ &TC$\downarrow$& {} &    FID$\downarrow$ &FVD$\downarrow$&LPIPS$\downarrow$&PSNR$\uparrow$& SSIM$\uparrow$& EDMD$\downarrow$ &TC$\downarrow$\\
 \midrule
 ACOF (prev)& 77.9654& 690.6636& 0.3671& 10.8886& 0.2634& 13.2373& 2.6077& & 91.6352& 588.8971& 0.3772& 10.7118& 0.2788& 15.2450&3.0174\\
 ACOF (first)& 25.2291& 280.1266& 0.1310& 19.6044& 0.8013& 7.8377 &1.2265& & 23.4283& 224.1050& 0.1097& 21.2396& 0.8208&8.6646 &1.2601\\
 \midrule
 TCVC (prev)& 77.6475& 779.2147& 0.3792& 11.7784& 0.4752& 11.3185& 1.8821& & 91.7752& 711.6013& 0.4191& 11.0152& 0.4570& 13.5151&2.0850\\
 TCVC (first)& 23.4118& 217.4509& 0.1419& 17.9502&  0.7587& 6.4081 &1.3628& & 26.6516& 224.9740&  0.1531& 17.7315& 0.7473&7.3664 &1.4165\\
 \midrule
 CNet + Refonly & 16.9340&  170.2000& 0.1114&  18.7949& 0.7844& 7.1903 &1.4581& & 17.1442& 135.8851& 0.0879& 20.6059& 0.8043&8.1612 &1.3272\\
 \midrule
 EISAI + CNet& 18.9262& 227.6656& 0.1413& 18.1603& 0.7314& 8.2527& 1.2952& & 19.0419& 190.6165& 0.1099& 20.0647& 0.7810& 9.5414&1.2306\\
 SEINE + CNet& 23.0797& 836.4528& 0.1655& 17.2425& 0.6504& 8.6777& 2.3644& & 30.5665& 820.9603& 0.1464& 18.2319& 0.6940& 10.6964&2.2858\\
 \midrule
 Ours& \textbf{8.8423}& \textbf{40.2711}& \textbf{0.0560}& \textbf{24.5489}& \textbf{0.8790}& \textbf{4.3386} &\textbf{1.0784}& & \textbf{8.8038}& \textbf{32.6929}& \textbf{0.0399}& \textbf{26.7859}& \textbf{0.9182}&\textbf{6.4648} &\textbf{1.1025}\\
 \toprule
\end{tabular}
\end{center}
\end{table*}

\begin{table*}[!ht]
    \centering
    \begin{minipage}{0.48\textwidth}
        \vspace{-2pt}
        \setlength\tabcolsep{3.1pt} 
        \footnotesize
        \caption{Ablation study on model architecture and sequential sampling.}
        \label{tab:ablation}
        \begin{center}
        \begin{tabular}{cccccc}
            \toprule
            {} &  \multicolumn{5}{c}{Similar Testset}\\
            {} & FID$\downarrow$ &    FVD$\downarrow$&LPIPS$\downarrow$& EDMD$\downarrow$ &TC$\downarrow$\\
            \midrule
            Full method& 8.8423& \textbf{40.2711}& 0.0560& \textbf{4.3386}&\textbf{1.0784}\\
            \midrule
            $-$ Ref Attn& 9.6793& 41.2857& 0.0694& 5.0137&1.1563\\
            \midrule
            $-$ Sample Schemes& \textbf{8.6321}& 40.5472& \textbf{0.0523}&  4.4863&1.1479\\
            \midrule
            Prev Sample& 11.1873& 42.2713& 0.0925& 5.4924&1.2475\\
            \toprule
        \end{tabular}
        \end{center}
    \end{minipage}%
    \hspace{5pt}
    \begin{minipage}{0.48\textwidth}
        \setlength\tabcolsep{3.1pt} 
        \footnotesize
        \caption{\zt{Ablation study on number of overlapped frames. The last column indicates the inference time ratio compared with no overlapping ($o=0$).}}
        \label{tab:overlap}
        \begin{center}
        \begin{tabular}{ccccccc}
            \toprule
            {} &  \multicolumn{6}{c}{Similar Testset}\\
            {} & FID$\downarrow$ &    FVD$\downarrow$&LPIPS$\downarrow$& EDMD$\downarrow$ &TC$\downarrow$ &Inf. Time$\downarrow$\\
            \midrule
            $o=2$& \textbf{8.7594}& 50.8983& \textbf{0.0558}& 4.3143&1.0949 &\textbf{1.17}\\
            \midrule
            $o=4$ (ours)& 8.8423& \textbf{40.2711}& 0.0560& 4.3386&1.0784 &1.40\\
            \midrule
            $o=6$& 8.8582& 45.2426& 0.0565&  \textbf{4.2694}&\textbf{1.0737}&1.75\\
            \midrule
            $o=8$& 8.8227& 44.3100& 0.0567& 4.3265&1.0785 &2.33\\
            \midrule
            $o=10$& 8.8683& 40.8429& 0.0578& 4.3795&1.0745 &3.50\\
            \toprule
        \end{tabular}
        \end{center}
    \end{minipage}
\end{table*}

\vspace{3pt} \noindent \textbf{Quantitative Comparison. \;}
In this section, we compare our method with others in terms of frame and video quality, frame similarity, sketch alignment with ground-truth animations, and temporal consistency, as specified in \refsec{exp-setup}.
As demonstrated in \reftab{comparison}, our method significantly outperforms other approaches in all aspects, particularly excelling in video quality (FVD) and temporal consistency (TC). 
These results signify our method's ability to generate long, temporally consistent animations of superior quality compared to previous works.
Furthermore, our method demonstrates the capacity to generalize to animations with diverse styles from the training dataset, yielding comparable quantitative results.
Overall, we introduce new metrics (EDMD and TC) to effectively evaluate the sketch alignment and temporal consistency of generated animations, setting a new standard for reference-based lineart video colorization.

\begin{figure}[t]
    \centerline{ \includegraphics[width=1.0\linewidth]{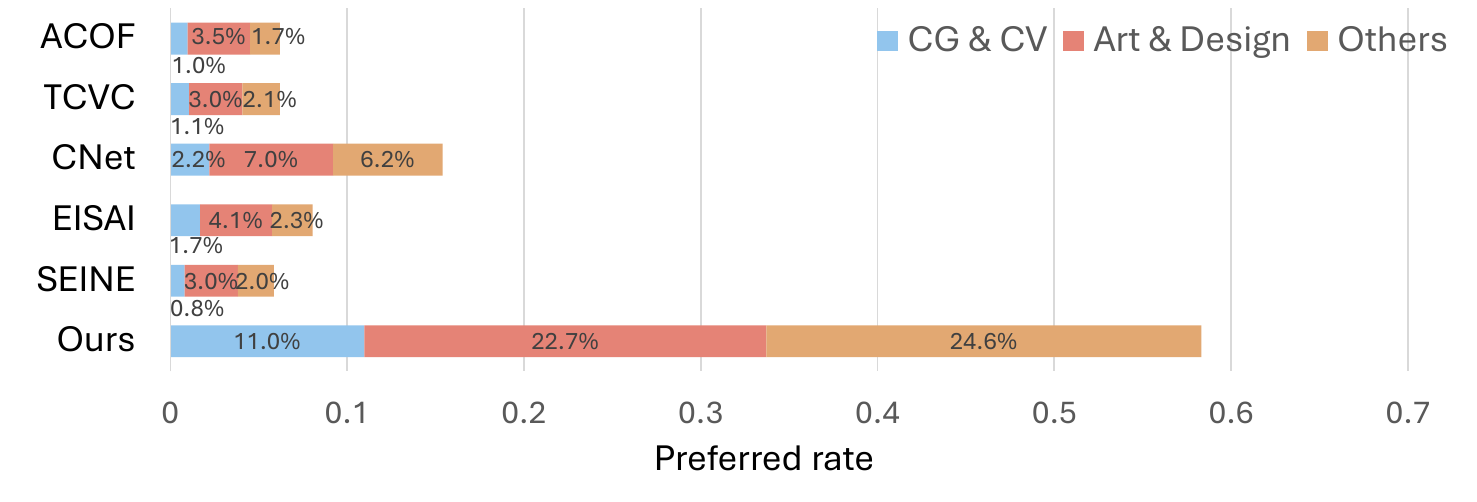} }
    \caption{\zt{Results of user study. Our method has a preferred rate of 58.3\% (62.0\%, 52.4\%, and 63.2\% for user group CG \& CV, Art \& Design, and others).}}
    \label{fig:user-study}
\end{figure}

\zt{
\subsection{User Study}
We conducted a user study to further evaluate our method's performance.
From a pool of 30 animations (15 from Similar Testset and 15 from General Testset), participants were presented with ground-truth animations and lineart sketches for reference, followed by animations generated by ACOF \cite{flow_anime_color}, TCVC \cite{temporal_video_color}, CNet+Refonly \cite{controlnet}, EISAI \cite{eisai}, SEINE \cite{seine}, and our method, in random orders.
\rv{Each user is required to answer 10 randomly selected questions from the pool by choosing the best animation considering all of the three aspects}: 1) similarity to the original, 2) alignment with lineart, and 3) overall quality.
\rv{Among 113 participants, 20 have worked or majored in areas related to Computer Graphics and Computer Vision (CG \& CV),} 49 are involved in Art \& Design, and 44 are from other fields.
As shown in \reffig{user-study}, our method garnered the highest preference rate of 58.3\%, with user groups from CG \& CV, Art \& Design, and other fields showing preference rates of 62.0\%, 52.4\%, and 63.2\%, respectively.
}


\subsection{Ablation Study}
\begin{figure*}[t]
    \centerline{\includegraphics[width=1.0\textwidth]{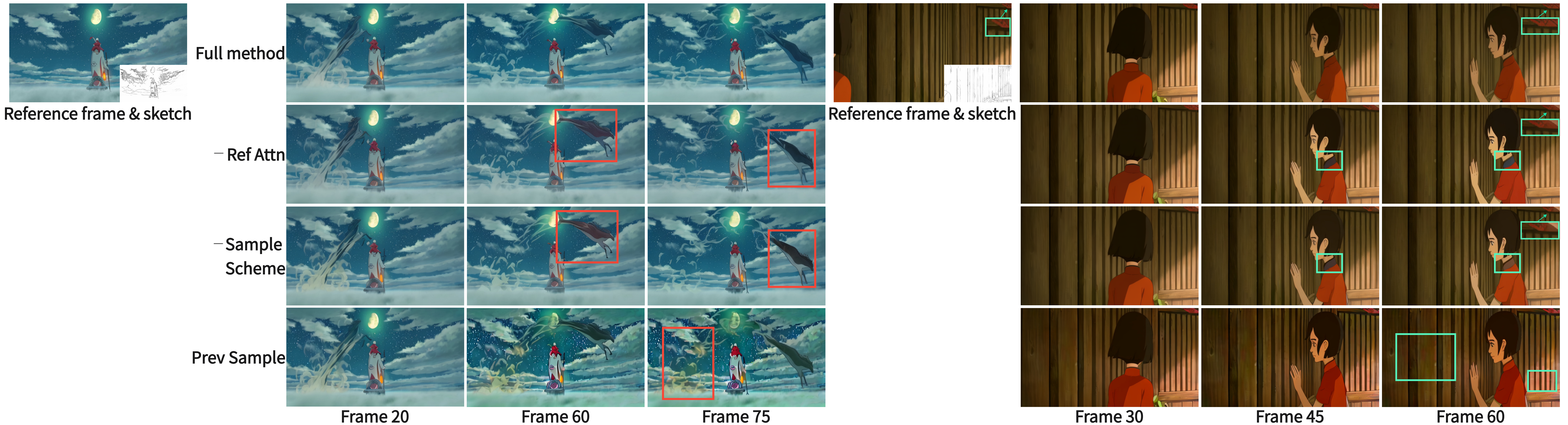}}
    \caption{\zt{Ablation study on model architecture and sampling scheme. All input frames are from \textit{Big Fish \& Begonia}.}}
    \label{fig:ablation}
\end{figure*}

\vspace{3pt} \noindent \textbf{Ablation on Model Architecture. \;}
To investigate the method's effect, we conducted an ablation study by removing the Reference Attention layers and re-training the network with the same hyper-parameters.
Quantitative results in \reftab{ablation} demonstrate a degradation across all metrics, indicating the model's diminished capacity to handle animations with large motions.
\zt{
Visually, as shown in \reffig{ablation}, the absence of Reference Attention leads to inconsistent and incorrect coloring of areas with significant motion, such as the black collar in the right example, as well as inconsistent colors of the deer in the left example.
}
These findings underscore the crucial role of Reference Attention in enhancing the model's ability to generate high-quality animations with large motions.


\begin{figure*}[t]
    \centerline{\includegraphics[width=1.0\textwidth]{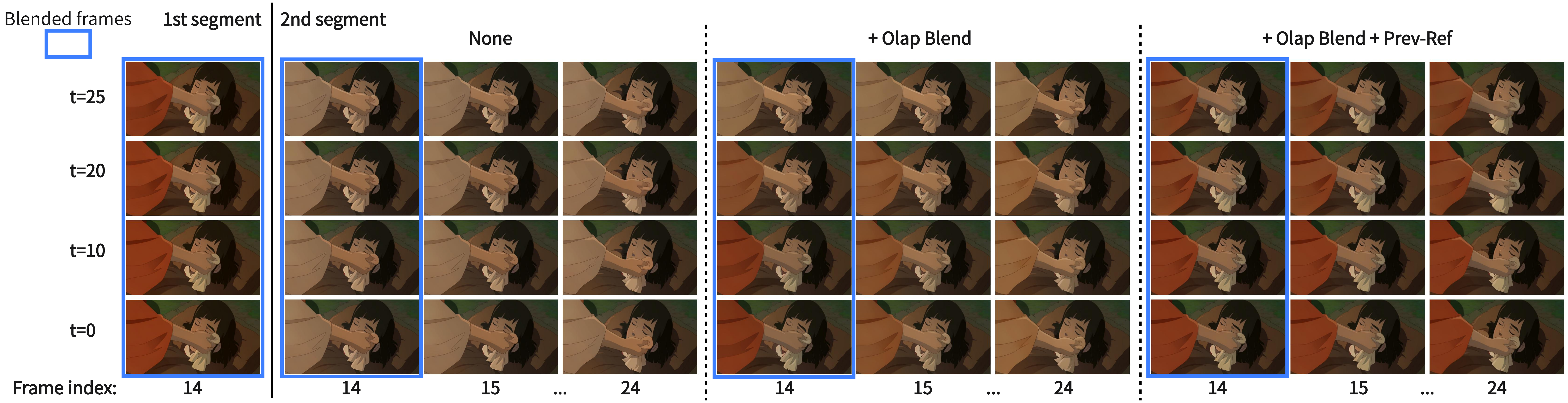}}
    \caption{\zt{Ablation study on sampling schemes including Overlapped Blending and Prev-Reference Attention. Input frame is from \textit{Big Fish \& Begonia}.}}
    \label{fig:ablation-sample}
\end{figure*}

\begin{figure}[t]
    \centerline{ \includegraphics[width=1.0\linewidth]{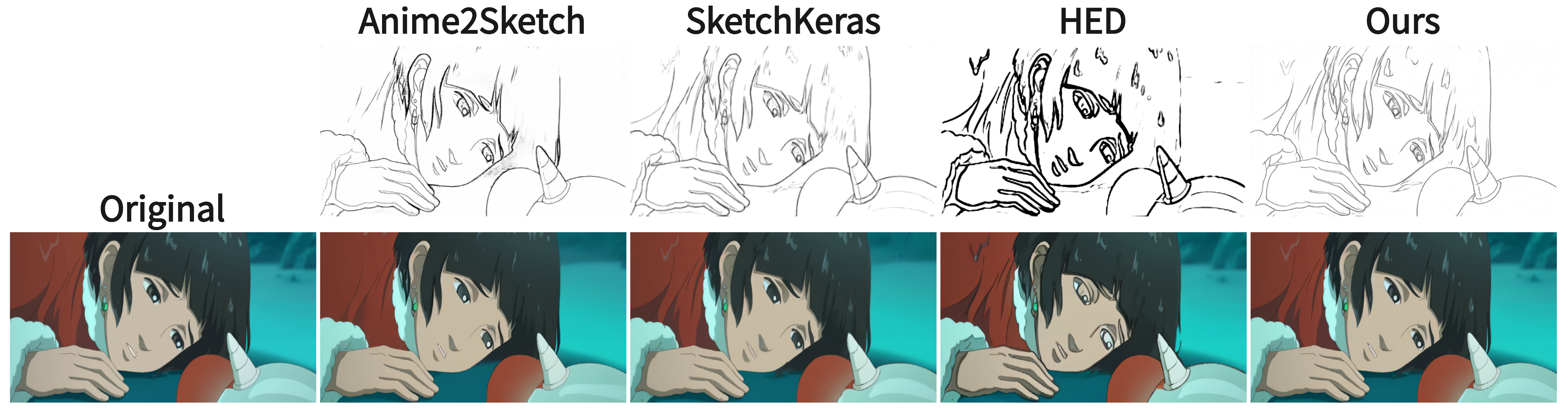} }
    \caption{\zt{Impact of different lineart extraction methods. Input frame is from \textit{Big Fish \& Begonia}.}}
    \label{fig:extraction}
\end{figure}

\vspace{3pt} \noindent \textbf{Ablation on Sequential Sampling Schemes. \;}
In \reftab{ablation}, we compare our method with two variants. Firstly, we remove all the sampling schemes mentioned in \refsec{sample-scheme} and sample the animations using only the first reference frame.
For Prev Sample, we further switch the reference to the generated frame from the previous segment.
Quantitatively, removing the sampling schemes leads to a reduction in temporal consistency (higher TC), while other metrics stay similar. For Prev Sample, all metrics degrade due to accumulated errors when using the previously generated frame as reference.

In the qualitative results depicted in \reffig{ablation}, our sampling schemes demonstrate superior ability to preserve temporal consistency.
For instance, the colors of \zt{the deer (left example)} and collar (right example) remain consistent with our full method, while inconsistencies arise in the results without sampling schemes.
Moreover, as the little plate in the left example (zoomed in the green frames) gradually reveals more area, our full method maintains a consistent red color akin to the reference image.
Compared to Prev Sample, our sampling schemes effectively mitigate accumulated artifacts. For instance, \zt{yellow region appears on the blue sky in the left example,} and the wall in the right example turns red with Prev Sample.
In summary, our sampling schemes can enhance long-term temporal consistency while simultaneously addressing the issue of accumulated artifacts.


In \reffig{ablation-sample}, we analyze the effects of the two schemes for sequential sampling, i.e., Overlapped Blending and Prev-Reference Attention.
We show the intermediate denoised outputs for $t = 25 \rightarrow 0$, where frame 14 of the $1^{st}$ segment overlaps with frame 14 in the $2^{nd}$ segment.
For the results without both modules, we note that the content (i.e. the red sleeve) of frame 14 cannot propagate from the $1^{st}$ to the $2^{nd}$ segment, leading to inconsistent brown sleeve in the newly sampled frame 14.
\zt{Upon integrating Overlapped Blending, the red sleeve in frame 14 in the $1^{st}$ segment can be inherited by the $2^{nd}$ segment.
Finally, with the incorporation of Prev-Reference Attention, the red color of the sleeve successfully propagates to frame 24, resulting in a temporally consistent animation.}


\begin{figure*}[t]
    \centerline{\includegraphics[width=1.0\textwidth]{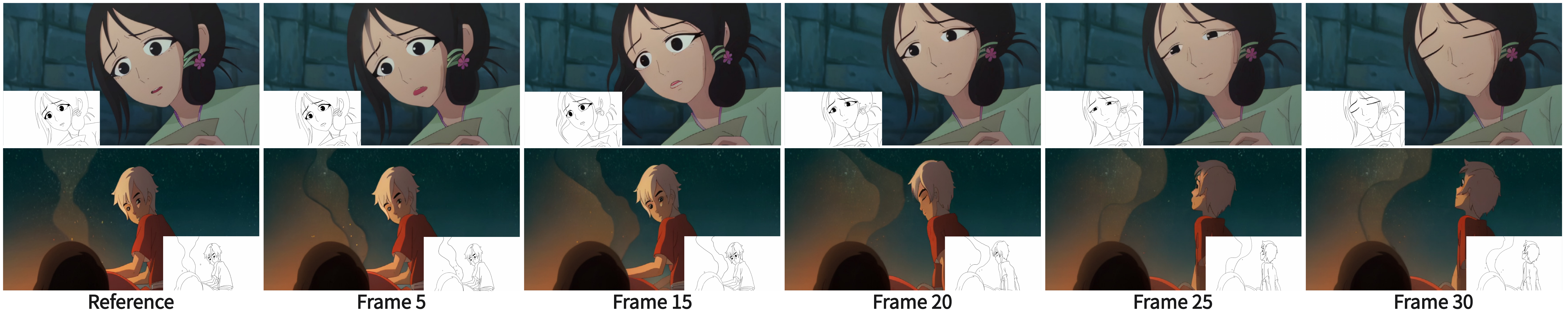}}
    \caption{Colorized results with hand-drawn sketches. Input frames: $1^{st}$ row is from \textit{Mr. Miao}, $2^{nd}$ row is from \textit{Big Fish \& Begonia}.}
    \label{fig:hand-drawn}
\end{figure*}

\begin{figure}[t]
    \centerline{ \includegraphics[width=1.0\linewidth]{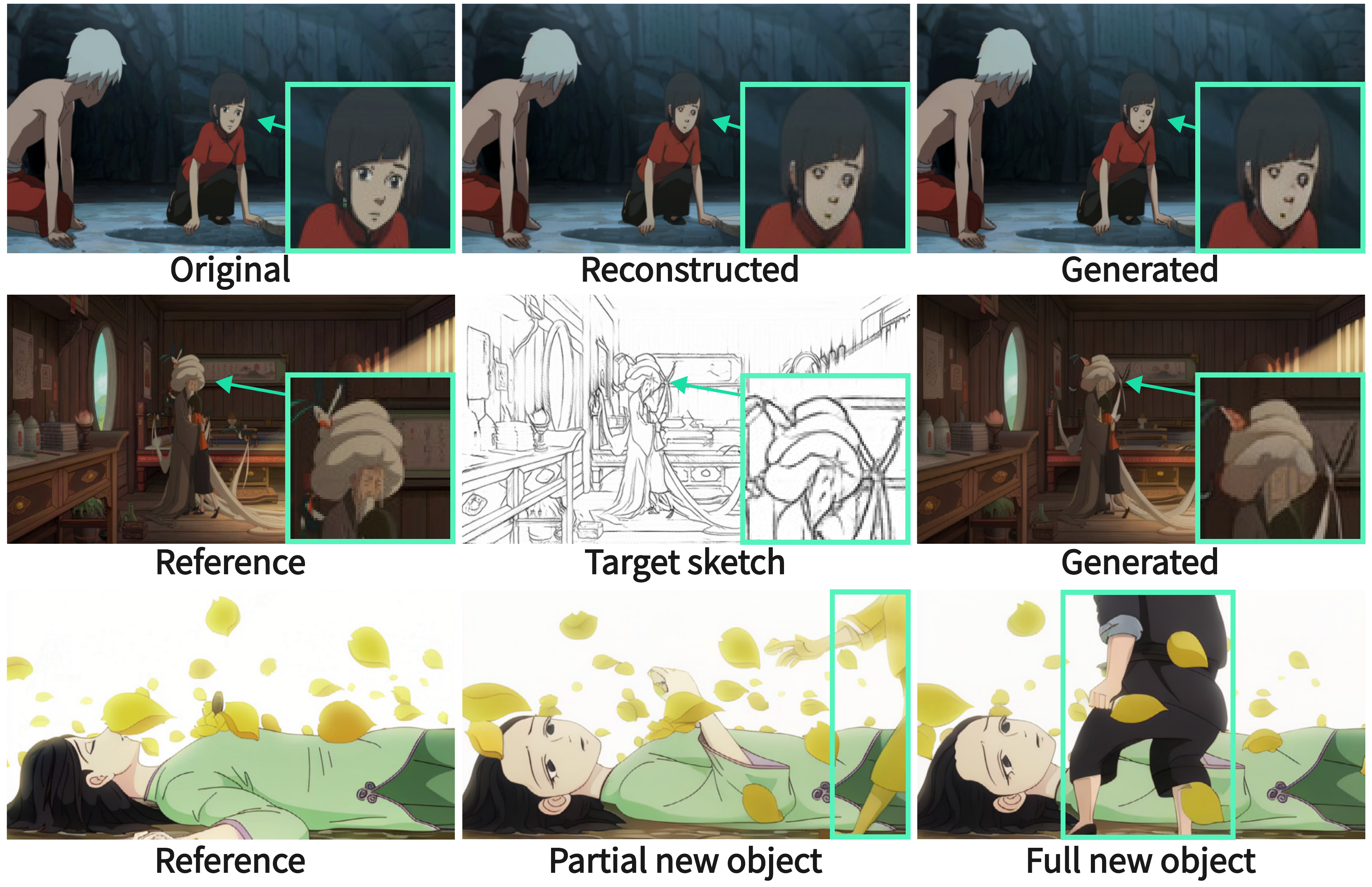} }
    \caption{\zt{Limitations. $1^{st}$ row: artifacts due to VAE. $2^{nd}$ row: artifacts due to coarseness in lineart sketches. $3^{rd}$ row: artifacts due to partial new object. Input frames: $1^{st}$ and $2^{nd}$ rows are from \textit{Big Fish \& Begonia}, $3^{rd}$ row is from \textit{Mr. Miao}.}}
    \label{fig:limitation}
\end{figure}

\zt{
\vspace{3pt} \noindent \textbf{Ablation on Number of Overlapped Frames. \;}
As discussed in \refsec{sample-scheme}, we divide the video sequences into segments with $o$ overlapped frames for sequential sampling.
In \reftab{overlap}, we analyze different numbers of overlapped frames.
We observe that increasing $o$ from $2$ to $4$ significantly improves video quality (FVD) and temporal consistency (TC), while other image quality metrics remain almost unchanged.
However, further increasing the number of overlapped frames does not yield significant improvements and even causes a decline in FVD for $o=6$ and $o=8$, while also slowing down inference speed.
Therefore, to balance the inference quality and speed, a setting of $o=4$ overlapped frames is the optimal choice.
}

\zt{
\vspace{-4pt}
\subsection{Impact of Different Lineart Extraction Methods}
To assess the impact of different lineart extraction methods, we apply our model, which was trained using linearts extracted from \cite{draw_sketch}, to colorize linearts generated by various extraction methods, including Anime2Sketch \cite{anime2sketch}, SketchKeras \cite{sketchkeras}, HED \cite{hed} (using combined features from all layers), and the method employed in our training dataset. 
As illustrated in \reffig{extraction}, our model is capable of producing results of similar quality for linearts extracted by Anime2Sketch and SketchKeras, despite the differences in style and detail from the linearts used in our training, demonstrating the generalizability of our method. 
However, when applied to linearts that are excessively thick, as in HED, our model tends to generate animations with thick and blurred lines.
This issue could be solved by augmenting the training data with linearts of different thicknesses.
}

\subsection{Application to Hand-drawn Linearts}
To verify the practical applicability of our method, we engaged students specializing in painting to create hand-drawn lineart sketches using graphic tablets.
Then, we utilized our method to colorize these hand-drawn linearts.
As shown in \reffig{hand-drawn}, our model, originally trained with automatically generated sketches, seamlessly accommodates the hand-drawn lineart sketches.

\zt{
\subsection{Limitations}
Despite our method's effectiveness, it has two limitations. 
Firstly, our method may suffer artifacts in tiny details, due to reconstruction loss of VAE and coarseness in the input sketches. As shown in \reffig{limitation}, the details of the girl's face are lost in $1^{st}$ row due to reconstruction loss, and the elder's face is blurred due to the coarseness of the sketches in $2^{nd}$ row.
Finetuning the original VAE to suit cartoon image domains and the resolution used in our framework and \rv{employing data augmentation} on the training sketches could mitigate this issue.
Another limitation is the potential for inaccurate colorization of partially visible new objects. 
As in the $3^{rd}$ row, when only a portion of a new character enters the scene, the body is incorrectly colorized as the color of nearby petals.
The colorization only becomes accurate when the character's full body is visible.
Modifying our video clipping algorithm to include more scene change cases that involve more new objects during training, may help our model handle such scenarios.
}

\section{Conclusions}
\label{sec:conclusion}

In summary, we introduce the first video diffusion framework for reference-based lineart video colorization, addressing the limitations of previous methods.
Our approach can produce long temporal-consistent animations of superior quality, by leveraging a pretrained video diffusion model.
To adapt the pretrained SVD to our task, we introduce sketch-guided ControlNet for controllable video generation, and Reference Attention, enabling the model to handle expansive motions.
Furthermore, our novel sequential sampling, including Overlapped Blending and Prev-Reference Attention, extends the model's capability to generate long animations while preserving temporal consistency.
Our experiments validate the efficacy of our method, showcasing its ability to produce high-quality animations with large motions, which is not achieved by previous works.

\zt{
As our framework is generic, it can be applied to other modalities, e.g., edge, depth, and normal map. In future works, we may extend our method to generate realistic videos guided by other modalities or even multi-modality. 
The performance of realistic video generation could be further improved by using large-scale realistic video datasets and leveraging the fact that SVD is also pretrained on similar videos.
}

\begin{acks}
We thank the anonymous reviewers for helping us to improve this paper. 
We also thank Bi An Tian (Beijing) Culture Co., Ltd., and Gu Dong Animation Studio for approving us to use their animation content.
The work described in this paper was fully supported by a GRF grant from the Research Grants Council (RGC) of the Hong Kong Special Administrative Region, China [Project No. CityU 11216122].
\end{acks}

\bibliographystyle{ACM-Reference-Format}
\bibliography{reference}




\cmt{
\begin{figure*}[!htb]
    \centering
    \begin{minipage}{0.48\textwidth}
        \vspace{101pt}
        \centering
        \includegraphics[width=1.0\linewidth]{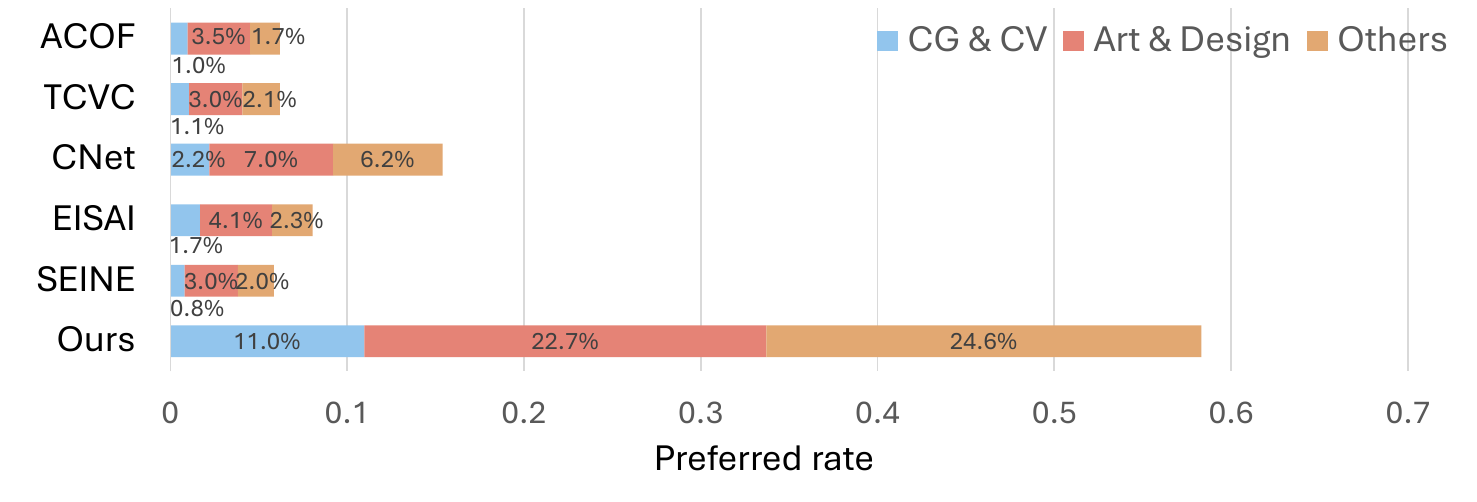}
        \caption{Results of user study. Our method has a preferred rate of 71.5\% (71.0\%, 74.2\%, and 70.6\% for user group CG \& CV, Art \& Design, and others).}
        \label{fig:user-study}
    \end{minipage}%
    \hspace{5pt}
    \begin{minipage}{0.48\textwidth}
        \centering
        \includegraphics[width=1.0\linewidth]{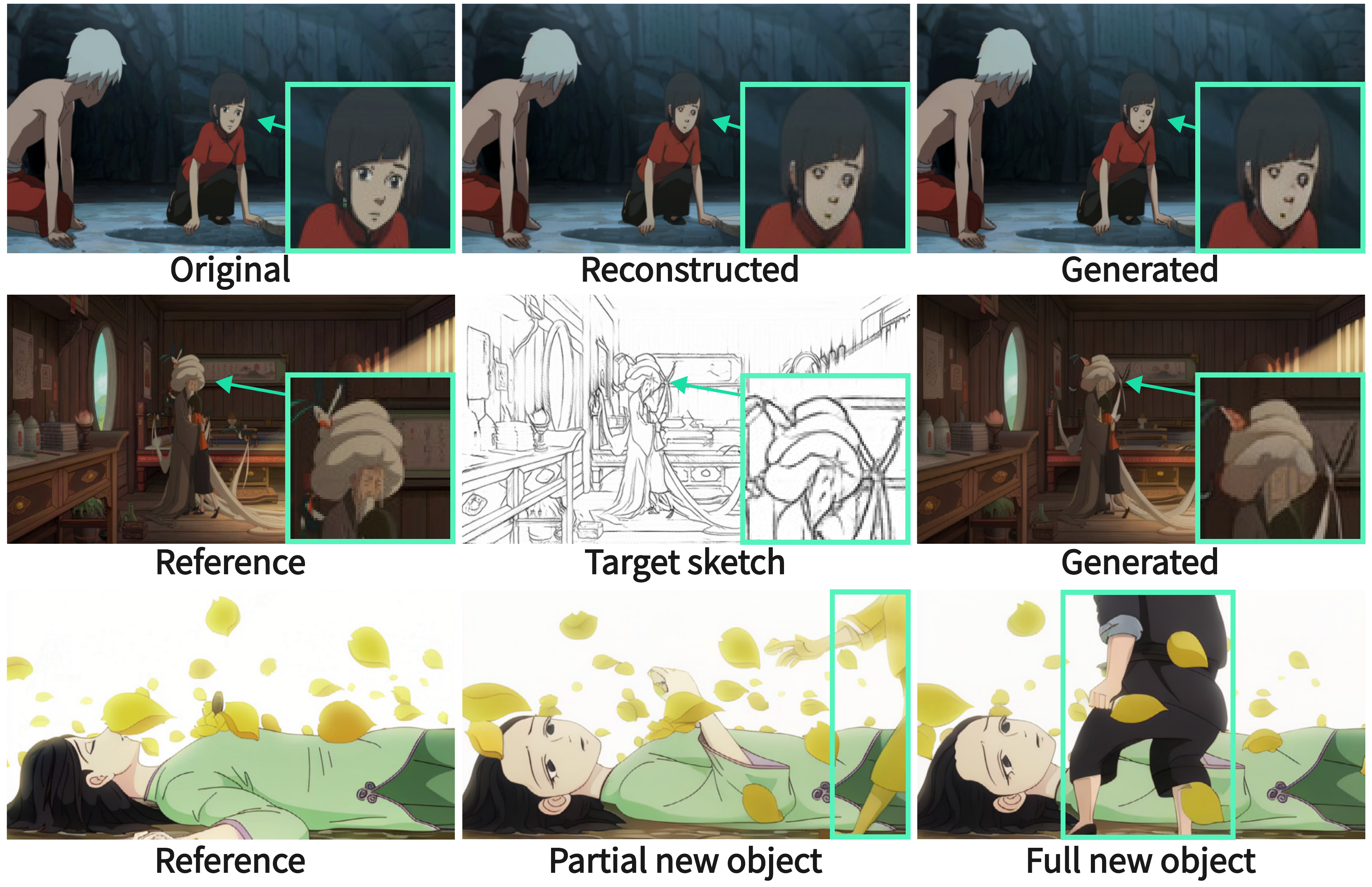}
        \caption{Limitations. $1^{st}$ row: artifacts due to VAE. $2^{nd}$ row: artifacts due to coarseness in lineart sketches. $3^{rd}$ row: artifacts due to scene change.}
        \label{fig:limitation}
    \end{minipage}
\end{figure*}
}

\end{document}